\documentclass{article}

\usepackage{times}
\usepackage[T1]{fontenc}
\usepackage{amsmath}
\usepackage{amssymb}
\usepackage{url}
\usepackage{graphicx} 
\usepackage{algpseudocode}
\usepackage{algorithmicx}
\usepackage{textcomp}
\usepackage{wrapfig}
\usepackage{tikz}
\usepackage{verbatim}
\usepackage[english]{babel}

\usetikzlibrary{arrows,shapes,backgrounds}

\newcommand{\ptolemy}{\textsc{Ptolemy II}}

\newcommand{\jena}{\textsc{Jena}}
\newcommand{\protege}{\textsc{prot\'eg\'e}}

\newcommand{\pellet}{\textsc{Pellet}}
\newcommand{\arq}{\textsc{ARQ}}
\newcommand{\quest}{\textsc{Quest}}

\newcommand{\ailift}{\textsc{AiLift}}
\newcommand{\liftcreate}{\textsc{LiftCreate}}

\title{Ontologies in System Engineering: a Field Report}

\author{%
  Marco Menapace and Armando Tacchella\\
  \small DIBRIS - Universit\`a degli Studi di Genova - Viale F. Causa 13, 16145 Genova\\
  \small \url{marco.menapace@edu.unige.it} - \url{armando.tacchella@unige.it}
}

\begin{document}

\maketitle

\begin{abstract}

  In recent years ontologies enjoyed a growing popularity outside
  specialized AI communities. System engineering is no exception to
  this trend, with ontologies being proposed as a basis for several
  tasks in complex industrial implements, including system design, monitoring 
  and diagnosis. In this paper, we consider four different
  contributions to system engineering wherein ontologies are 
  instrumental to provide enhancements over traditional ad-hoc 
  techniques. For each application, we briefly report the
  methodologies, the tools and the results obtained with the goal to
  provide an assessment of merits and limits of ontologies
  in such domains.   

\end{abstract}

\section{Introduction}
\label{sec:intro}

Ontologies are witnessing an increasing popularity outside specialized AI
communities. While this is mostly due to Semantic Web
applications~\cite{ber01}, we must also credit their ability to
cope with taxonomies and part-whole relationships, to
handle  heterogeneous attributes, and their provision for various
automated reasoning services --- see, e.g.,~\cite{staab2013handbook}.  
These features have been recognized since long time in system engineering,
the community encompassing all areas of research devoted to design,
implementation, monitoring and diagnosis of technical processes.
For instance, in the operations and maintenance sub-community, the use
of ontologies is explicitly advocated\footnote{See, e.g., the MIMOSA
open standard architecture at \url{www.mimosa.org}.}.  
Also standards like ISO 13374
(\textit{Condition monitoring and diagnostics of 
machines -- Data processing, communication and presentation}) suggest
the use of ontologies for several tasks, mostly related to data
conceptualization. However, the adoption of ontologies
faces some challenges, mostly due to speed and reliability constraints
imposed by industrial settings.  

Here we investigate this issue by considering four contributions of
ours to application domains wherein ontologies provide key
capabilities in system engineering. The first case study is
about an on-board rolling-stock condition analyzer, i.e., a
system to perform fault detection and
classification~\cite{AmbrosiGT09}. The second one is about monitoring 
an intermodal logistic system~\cite{CasuCT13}. The third one is about an
ontology-based framework to generate diagnostic-decision support
systems~\cite{CicalaLOT16}. Finally, a fourth case study is an 
application to computer-automated design of elevator
systems.
In the following, we briefly introduce each case study, giving details
about its context, underlying motivation and intended objectives. The 
ultimate goal of the paper is to discuss and compare the results
obtained to assess the effectiveness of ontologies in such
application domains.  

\noindent
\textit{Ontologies for condition analysis.}
We introduced an ontology-based condition analyzer
(CA)~\cite{AmbrosiGT09} in the context of the EU project  
Integrail\footnote{More details about Integrail
at \url{http://www.integrail.eu/}.}.      
Our CA collects signals from control logic installed on locomotives,
and it leverages an ontology to correlate
observed data, symptoms and faults. 
The CA must mate two competing
needs: $(i)$ railway regulations
require hardware which is highly reliable, and whose performances are
thus far even from desktop workstations; $(ii)$
ontology-related tools, e.g., description logic reasoners,
have relatively large memory, processor and storage footprints.
In this experience, the main 
goal was thus to check whether reasoning with ontologies can provide
useful diagnostic feedback in a resource-restricted scenario. 

\noindent
\textit{Ontologies for system monitoring.}
In~\cite{CasuCT13} we provided strong evidence of practical uses
for ontologies in complex systems engineering by implementing
a monitor for \emph{Intermodal Logistics Systems} (ILSs), i.e.,
systems supporting the movement of containerized goods. In
particular, we considered combination of rail and road transport,
where rail transport is provided by short-distance shuttle trains,
and network coverage is
achieved through connections at specialized terminals. In this
experience, the main goal was to gather data about terminal operations
and compute global performances indicators, where access to data is
mediated by an ontology --- ontology-based data access
(OBDA)~\cite{cal05}. Here, unlike the CA case study, 
the ability to handle large amount of data is crucial, but reasoning
is limited to SPARQL query answering.

\noindent
\textit{Ontologies for diagnostic support system generation.}
Diagnostic Decision Support Systems (DDSSs) help humans to 
infer the health status of physical systems. In~\cite{CicalaLOT16} we
introduced DiSeGnO --- for ``Diagnostic Server Generation
through Ontology'' ---  to generate customized DDSSs.
As in the ILS monitoring case study, since it is
expected that large quantities of data should be handled, 
the ontology language is
restricted to those designed for tractable reasoning --- see,
e.g.,~\cite{cal05}. In this case, ontology-based  
reasoning is not leveraged, as DiSeGnO generates relational databases
from the domain ontology and then computes diagnostic rules
with \ptolemy{}~\cite{Ptolemy}, an open-source software simulating
actor-based models.  

\noindent
\textit{Ontologies for computer-automated design.}
As mentioned in~\cite{ByeOPHS16}, the first scientific report of
intelligent computer-automated design (CautoD) is the paper by
Kamentsky and Liu~\cite{KamentskyL63}, who created a computer program
for designing character-recognition logic circuits satisfying given
hardware  constraints. In mechanical design --- see,
e.g.,~\cite{rao2012mechanical} --- the term usually refers to
techniques that mitigate the effort in exploring alternative
solutions for structural implements. In our \liftcreate{}
CautoD program for elevator systems\footnote{Part of the \ailift{}
software suite \url{www.ailift.it}.}, ontologies  
support intelligent design creation and optimization by managing
detailed part-whole taxonomies, wherein different relations among
components can be expressed. This case study provides thus yet another
application of ontologies, mostly oriented to intelligent computation
and data persistency.

Overall, the case studies considered witness the great flexibility
that ontologies provide in handling diverse application scenarios,
from condition analysis of locomotives, to automated design of
elevators, considering both cases wherein they provide the basis for
logic reasoning services, or just advanced data-modeling capabilities.  
The rest of the paper is structured as follows. In
Sections~\ref{sec:condition}, \ref{sec:ilog}, \ref{sec:ondaBrief}
and \ref{sec:elevator} we sketch the design, the implementation and
the results obtained in the case studies described above.
Section~\ref{sec:concl} concludes the paper by summarizing the results
and providing some discussion thereof.

\section{Rolling stock condition analysis}
\label{sec:condition}
\begin{figure}[t!]
\begin{center}
  \includegraphics[scale=0.27]{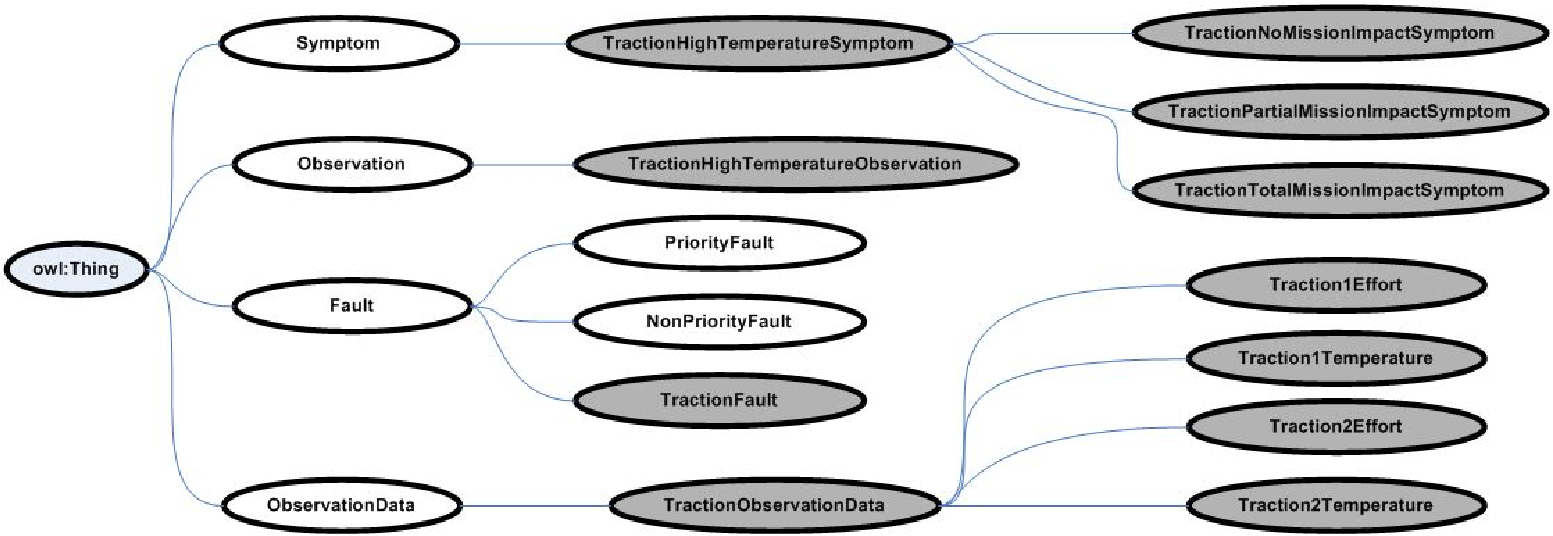}
  \caption{\label{fig:e414ont}A portion of the E414 ontology regarding
    traction faults. Concepts are nodes and
    object properties are edges: white nodes are SP3A
    concepts, grey ones are E414-specific concepts.}
\end{center}
\end{figure}

The CA prototype described in~\cite{AmbrosiGT09} focuses on 
fault detection on Trenitalia E414 locomotive.
The main task 
of the CA is to perform fault classification according to 
priority for maintenance, and impact on mission-related and
safety-related aspects.
Here, we focus on traction groups as an example of
subsystem that can generate a faulty condition.
Our ontology for the E414 locomotive is written in OWL 2 language
and it builds on the SP3A core ontology 
--- see~\cite{AmbrosiGT09} for details. 
In particular, the E414 ontology leverages the SP3A concepts of {\sc
  ObservationData}, i.e., process variables, and {\sc Observation},
i.e., sequences of observation data from which individuals of class
{\sc Symptom} and {\sc Fault} arise. {\sc Symptom} individuals
are related to {\sc Observation} individuals via the {\sc
  refersToObservation} property and to {\sc Fault} individuals via the
{\sc refersToFault} property. {\sc Fault} is a concept whose
individuals are defined in terms of the necessary {\sc hasSymptom}
relationship with {\sc Symptom} individuals.
Two subclasses of {\sc Fault} are defined: {\sc
  PriorityFault} and {\sc   NonPriorityFault}, with obvious
meaning. 
In Figure~\ref{fig:e414ont} we show a portion of the E414 ontology
related to  traction faults, where concepts have been specialized 
in subclasses whose individuals correspond to actual signals and
subsystems. Fault and symptom classification is obtained by
a Description Logic (DL) reasoner considering the patterns 
observed. For instance, in the case of {\sc
  TractionHighTemperatureObservation}, three ranges of temperatures
are defined that correspond to ``interesting'' patterns: from 70 to 80
degrees, from 80 to 130 degrees, and over 130 degrees. It is 
postulated that observations falling in the second and in the
third ranges are to be considered mission critical, while the ones in
the first category are only maintenance critical. 



A detailed description of the CA architecture can be found
in~\cite{AmbrosiGT09}. Here we provide some intuition on how the
analyzer works considering high temperatures in the traction groups. 
When the temperature of a group is higher than 70 
degrees for at least 3 consecutive samples read from the field bus,
the CA starts tracking a potential anomalous pattern.
Once such a pattern is detected, the corresponding
individuals in the classes \textsc{TractionObservationData} 
and \textsc{TractionHighTemperatureObservation}
are recorded. {\sc Symptom} individuals
are built along with all the properties required by the ontology
specification. For example, if an observation of the class
\textsc{TractionHighTemperatureObservation} has been created,   
a specific individual 
\textsc{TractionHighTemperatureObservation} is related to  a new
\textsc{Symptom} individual by the
\textsc{refersToObservation} property.
{\sc Fault} individuals for each {\sc Symptom} individual are created 
together with the {\sc causedBySymptom} property.
{\sc Fault} as well as {\sc Symptom} individuals are built of generic
type, leaving their classification to the DL reasoner. Once
the classification is done, the CA publishes the results, transmitting
them to external agents. As an example, let us assume that $i$ is an
individual of the class \textsc{TractionHighTemperatureObservation}
whose property \textsc{isAt} is set to the constant
\textsc{\_130degrees}, $s$ is the \textsc{Symptom} individual related
to $i$, and $f$ is the \textsc{Fault} individual related to $s$. 
The E414 ontology postulates that all symptoms such that the
corresponding observation is an instance of
\textsc{TractionHighTemperatureObservation} related by \textsc{isAt}
to the constant \textsc{\_130degrees}
are also an instance of \textsc{TractionTotalMissionImpactSymptom},
which is a subclass of \textsc{Symptom}. Therefore, a reasoner 
can infer that $s$ belongs to 
\textsc{MissionRelatedSymptom}.


\begin{table}[t!]
\caption{\label{tab:results} Results with 
  (a) lazy and (b) eager implementations of the CA.}
\begin{center}
\tiny
    \begin{tabular}{ | c | c | c | c | }
    \hline			
      Scenario & Memory Consumption [MB] & CPU Time [ms]  & Amortized CPU Time [ms] \\
    \hline
      1a & 38 & 90 & ND \\
      2a & 74 & 25373 & 25373 \\
      3a & 106 & 1053656 & 210731 \\
      4a & OUT OF MEMORY & 3253637 & 191390 \\
    \hline
      1b & 37 & 90 & ND \\
      2b & 72 & 21506 & 21506 \\
      3b & 104 & 86938 & 17387 \\
      4b & 105 & 279523 & 16442 \\
    \hline
    \end{tabular}
\end{center}
\end{table}

Out of the three sets of experiments performed in~\cite{AmbrosiGT09},
we report just those to ensure that the CA
implementation fits the constraints. To this end, we ran several tests
using different fault scenarios\footnote{Sets  
  of multidimensional time series (3600 samples at 1Hz) corresponding
  to 52  process variables are generated.
  Simulations run on EN50155-compliant embedded devices 
  with 1GHz Socket 370 FC-PGA Celeron Processor with 256MB 
  of main memory and a 1GB SSD running Linux Blue Cat (kernel 2.6) and
  Sun Java Virtual Machine implementation (JRE 1.6). The DL reasoner is \pellet{}~\cite{sirin07}.}.
Table~\ref{tab:results} shows the results obtained by running the CA
on four different scenarios --- the first includes no
fault, the second includes only one fault, the third includes five
contemporary faults, and the last 17 contemporary faults --- 
using two different configurations.
Configuration (a) is ``lazy'', i.e., it keeps all the individuals,
while configuration (b) is ``eager'', i.e., it  
deletes individuals as soon as possible.
As we can see in Table~\ref{tab:results}, the eager version results in
a great improvement over the lazy one, both in terms of memory
consumption and in terms of computation time.
In particular, in the second column of Table~\ref{tab:results} we can
notice that the eager version performs reasonably well, even in the
fourth test case (worst-case scenario). In the same scenario,
the lazy version exceeds the amount of available memory.
As we can see in the rightmost column of Table~\ref{tab:results}, the
amortized computation time over a single scenario
decreases with the number of concurrent
observations detected in the round.
Managing a round of samples without detected observations
takes only 90 ms, which leaves enough time for other activities, and
allows the CA to process all the incoming signals in due course.

\section{Monitoring of intermodal systems}
\label{sec:ilog}
\begin{figure}[t!]
\centerline{\scalebox{.26}{\includegraphics{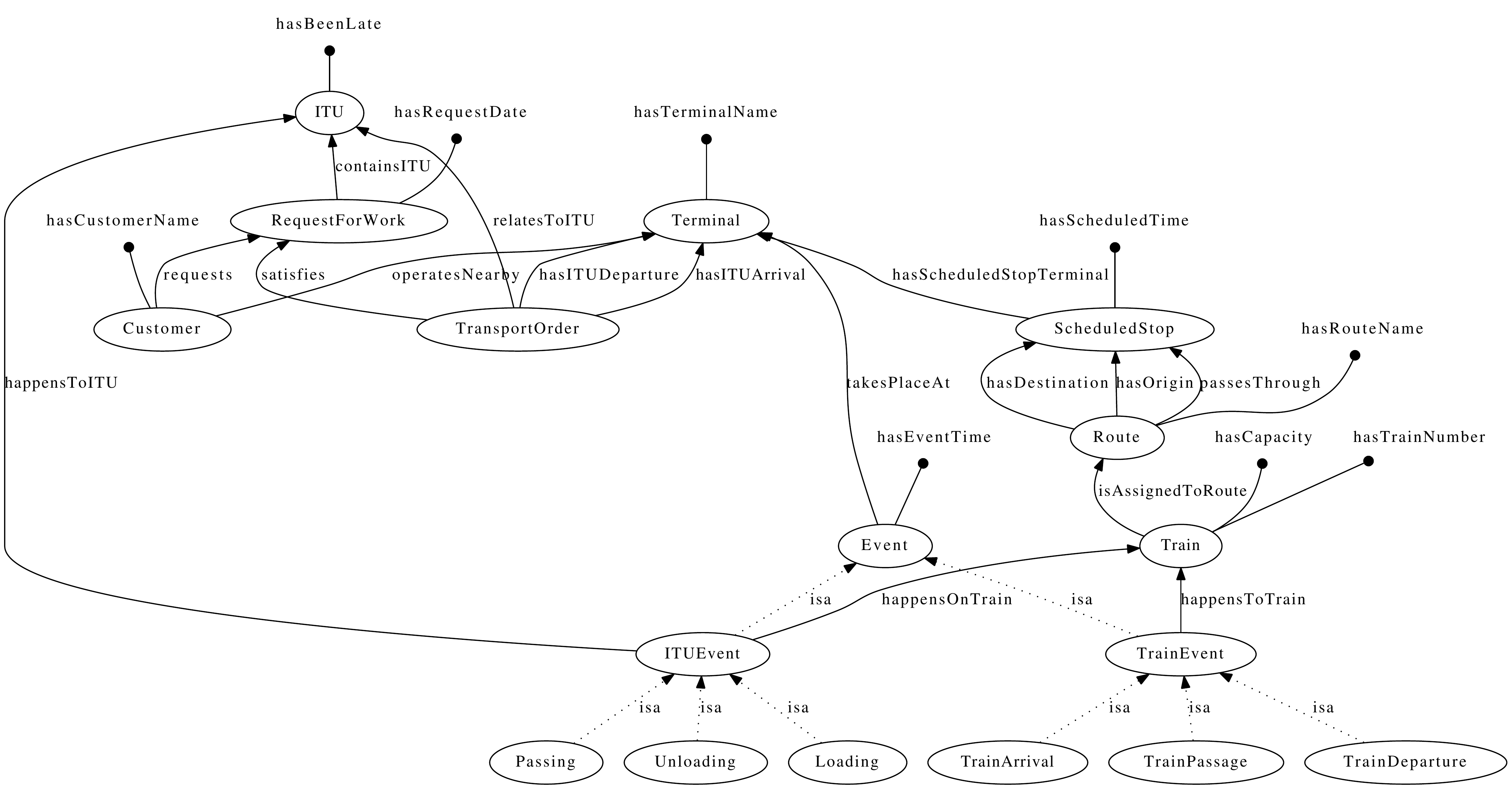}}}
\caption{\label{fig:iLog_dot} ILS ontology describing the design of the
  OBDA solution. Ellipses denote concepts with datatype properties;
  directed edges are object properties; dotted edges are concept
  inclusions.} 
\end{figure}

In~\cite{CasuCT13} we provided evidence that ontology-based data
access (OBDA)~\cite{cal05}  
is of practical use in the context of \emph{Intermodal Logistics
  Systems} (ILSs). The investigation focuses on the opportunity to build a monitoring information
system (MIS) using OBDA instead of relational databases
(RDBs). The application scenario is an ILS relying on a
logic akin to computer networks, i.e., frequent short-distance trains
with a fixed composition and a predefined daily schedule to cover some
geographical area. \emph{Intermodal Transport Units} (ITUs) enter
the network at some terminal and travel to their destination according
to a predefined route, usually boarding more than one train along the
way. Terminals collect ITUs from areas of approximately 150Km in
radius in order to minimize road transport. The MIS is a key
enabler to minimize delivery time, maximize rolling-stock and network
utilization and, ultimately, reduce the economic overhead of
transportation for the final customer. The main goal of the MIS is to
compute \emph{Key Performance Indicators} (KPIs) to perform tactical and 
strategical decision making about the network.

In Figure~\ref{fig:iLog_dot} we present a graphical outline of the
ontology at the heart of our OBDA solution to monitor the ILS.
The ontology --- ILS ontology in the following --- is compliant with
the OWL 2 QL profile described in the official W3C's recommendation
as \emph{``[the sub-language of OWL 2] aimed at applications that use 
  very large volumes of instance data, and where query answering is
  the most important reasoning task.''}. Given the ILS application
domain, OWL 2 QL guarantees that conjunctive query answering and
consistency checking can be implemented efficiently
with respect to the size of data and ontology, respectively. The
restrictions that OWL 2 QL implies did not hamper the modeling
accuracy of our ILS ontology. In Figure~\ref{fig:iLog_dot} we can
pinpoint classes related to freight forwarding such as 
\textbf{Customer}, i.e., companies forwarding their goods through the
network, \textbf{RequestForWork}, i.e., the main document witnessing
that a given customer has issued a request for transporting a number
of ITUs, \textbf{TransportOrder}, i.e., the ``bill of transit''
associated to each ITU, as well as entities related to physical
elements such as \textbf{ITU}, \textbf{Terminal} and \textbf{Train}. Also
``logical'' entities are modeled such as \textbf{Route}, i.e., a
sequence of terminals and railway connections serviced regularly by
one or more scheduled trains and \textbf{ScheduledStop}, i.e.,
terminals associated to a given route with a given
schedule. \textbf{Event} is the main monitoring entity, as the
calculation of most KPIs relies on the exact recording of events at
specific locations. 

\begin{figure}[t!]
\begin{center}
  \scalebox{.27}{\includegraphics{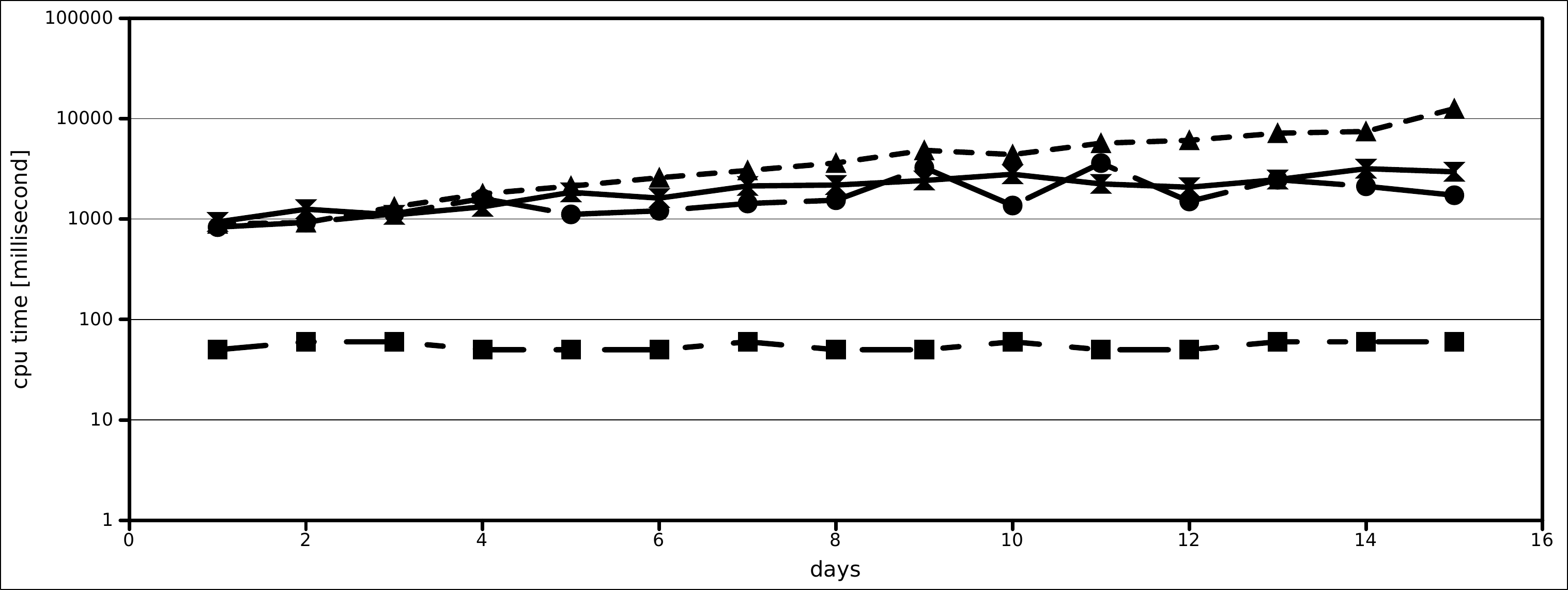}}
\end{center}
\caption{\label{fig:results} Computation time of a KPI with different
    query processors: SQL (square), \arq{} (circle), \pellet{}
    (hourglass), \quest{} (triangle). In each plot, the $x$ axis
    displays the number of simulation days from 1 to 15, the $y$ axis
    displays the CPU time (in milliseconds on a logarithmic
    scale).} 
\end{figure}

To assess OBDA performances, in~\cite{CasuCT13} we obtained  different
artificial utilization scenarios by changing 
the number of ITUs shipped daily from each terminal. Considering  
typical usage patterns, we postulated that a
provision of 10 to 50 ITUs is to be shipped daily from each terminal,
with 40 to 50 ITUS corresponding to a heavy utilization.
Scenarios are simulated for an increasing number of days to
evaluate scalability, and all of them share common settings as far as
number of train travels, number of cars per train, and timetabling are
concerned. Unexpected delays as well as the number of customers per
terminal follow a probabilistic model --- see~\cite{CasuCT13} for more
details. In Figure~\ref{fig:results} we display the results\footnote{
  All results are obtained on a family of identical
  Intel-based PCs, featuring a Core2Duo 2.13 GHz CPU, 4GB of RAM and
  running Ubuntu Linux 10.04 (64 bit edition).} obtained
in the case of an heavy utilization scenario to compute a specific
KPI, namely the average number of ITUs unloaded per hour.
The performance of four different query-answering systems
are reported: a SQL query on a native RDB implementation, and a SPARQL
query on the ontology store. The SPARQL query can be answered by three
different systems, namely \arq{} (the default query processor in the
\jena{} library), \pellet{} (the same DL reasoner that we consider in
Section~\ref{sec:condition}) and \quest{}~\cite{mur12}. The latter is the only
reasoner exploiting the fact that SPARQL queries can be compiled
on-the-fly into SQL queries for an equivalent RDB representation of
the ontology stored in the main memory. As we can observe in
Figure~\ref{fig:results}, OBDA-based solutions show higher overall
computation times than the RDB-based solution --- from 1 to 2 orders
of magnitude --- together with an apparently growing trend associated
to the time span of the simulation. However, as we have shown
in~\cite{CasuCT13}, a trend test performed on the results obtained
with the best OBDA solutions for various KPIs, displays no statistically
significant increase in the CPU time required to answer various
queries with respect to the number of days. Considering that for most
KPIs we can adopt an ``eager'' solution similar to that considered in
Section~\ref{fig:results}, we can conclude that OBDA is practically
feasible for monitoring medium-to-large scale systems.

\section{Diagnostic support systems generation}
\label{sec:ondaBrief}
\begin{figure}[!t]
\centering
\scalebox{0.34}{\includegraphics{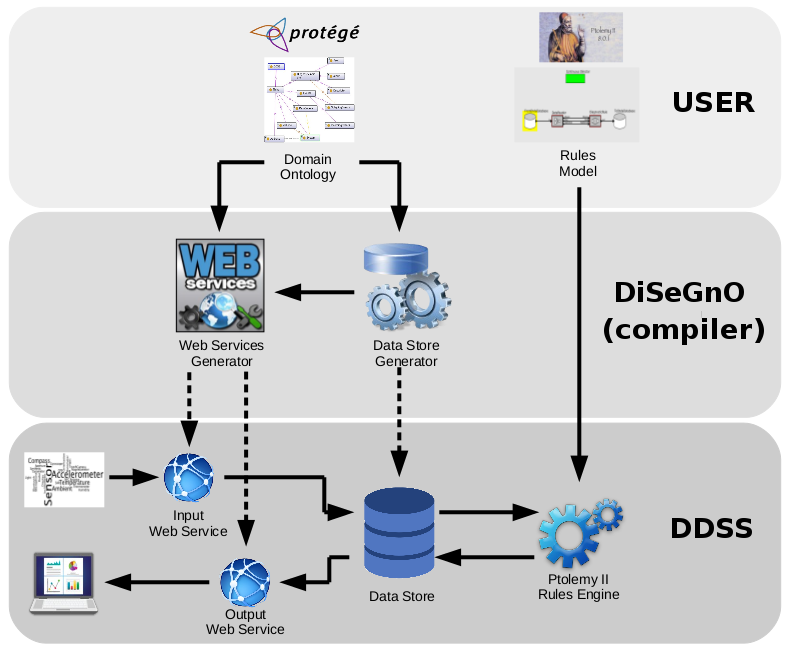}}
\caption{\label{fig:ondaModel}Functional architecture and work-flow of DiSeGnO framework.}
\end{figure}

In~\cite{CicalaLOT16} we introduced an approach to compile ontology-based
descriptions of equipment into diagnostic decision support systems
(DDSSs). The tool DiSeGnO, whose functional architecture and work-flow
is sketched in Figure~\ref{fig:ondaModel}, fulfills  this task in
three phases: in the \textsf{USER} phase, a domain ontology and
diagnostic rules model are designed by the user; in the
\textsf{DiSeGnO} phase, the system reads and analyzes the  ontology
and the rules to output the actual DDSS; in the \textsf{DDSS} phase,
input web services receive data from the observed physical system and
record them in the generated data  store. According to the ISO 13374-1
standard a DDSS consists of six modules of which DiSeGnO implements three:
\textit{Data Manipulation} to  perform signal analysis and 
compute meaningful descriptors, \textit{State Detection}
to check conformity to reference patterns, and \textit{Health
  Assessment} to diagnose faults and rate the current  
health of the equipment or process.
As shown in Figure~\ref{fig:ondaModel}, the ontology description is
created by a system architect in the \textsf{USER} phase. 
The ontology must be written using OWL 2 QL language\footnote{While
  this can be accomplished in several ways,  the tool
  \protege{}~\cite{gennari2003evolution} is suggested  because it 
is robust, easy to use, and it provides, either directly or through
plug-ins, several add-ons that facilitate ontology design and
testing.} as in the case study shown in Section~\ref{sec:ilog}. 
The diagnostic computation model must be a sound
actor diagram generated by \ptolemy{}~\cite{Ptolemy} which describes
the processing to be applied to incoming data in order to generated
diagnostic events --- here we focus on the ontology part, but more
details on the rule handling part can be found in~\cite{CicalaLOT16}.
The \textsf{DiSeGnO} phase contains the actual DDSS generation
system which consists of the \textsf{Data Store Generator}, i.e.,
a piece of software that creates a relational  database 
by mapping the domain ontology to suitable tables, 
and the \textsf{Web Services Generator}, i.e., a module  
that creates interface services for incoming and outgoing events.
Finally, in the \textsf{DDSS} phase, 
data is acquired and stored in the internal database,
the rules engine processes data and generates diagnostic events which
are then served to some application.  

\begin{figure}[t!]
\centering
\scalebox{0.25}{\includegraphics{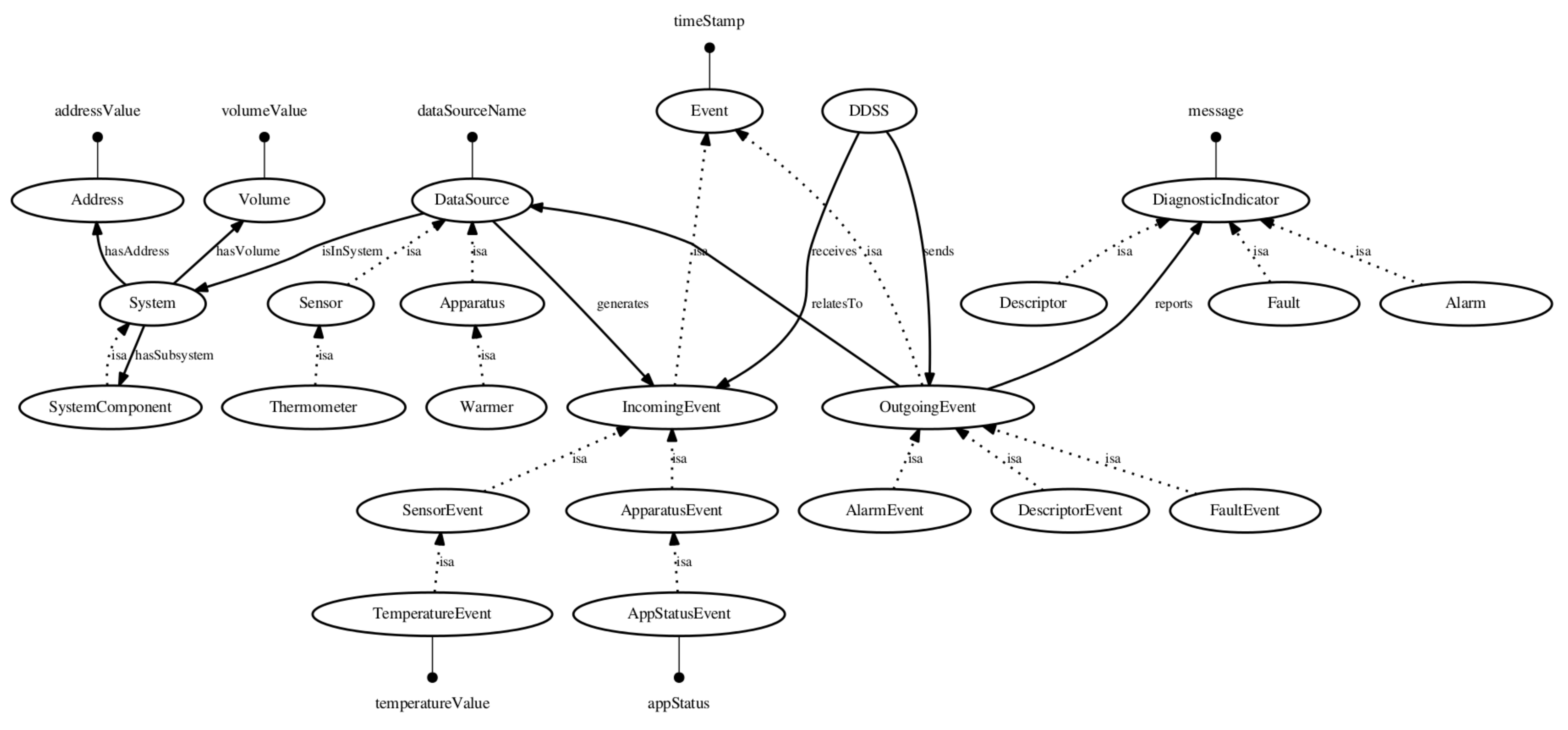}}
\caption{\label{fig:hvac_onto} Domain ontology for HVAC monitoring.
  Formalism is the same as in Figure~\ref{fig:iLog_dot}.}
\end{figure}

An example of a DiSeGno-compliant equipment description 
is shown in Figure~\ref{fig:hvac_onto}. The ontology is related to a 
Heating Ventilation and Air Conditioning (HVAC) appliance and it is
divided into  a \emph{static} and a \emph{dynamic} part. In the static
part, which is not updated while monitoring, the ontology contains a
description of the observed physical system. In the
HVAC ontology we have \textbf{System} and
\textbf{DataSource},  
related by the \textbf{isInSystem} property. \textbf{hasSubsystem}
relationship indicates that one \textbf{System} could be composed  by
one or more \textbf{SystemComponent} which are themselves subclasses
of  \textbf{System}. Finally, \textbf{DataSource} is the 
class of elements that can generate diagnostic-relevant information.   
The dynamic part describes \emph{events}, including both
the ones generated by the  observed system and its components, and
those output by the DDSS. An event is always associated to a
time-stamp and it can be either \emph{incoming} to the DDSS from the
observed system, or \emph{outgoing} from the DDSS\footnote{This
  distinction is fundamental, because DiSeGnO must know which 
events have to be associated with input and output web services,
respectively.}.
The main concepts in the dynamic part of the HVAC
ontology are \textbf{DDSS} which \textbf{receives} instances of
\textbf{IncomingEvent} and \textbf{sends} instances of
\textbf{OutgoingEvent}. Notice that \textbf{IncomingEvent} instances
are connected to \textbf{DataSource} instances by the role
\textbf{generates}, denoting that all incoming events
are generated by some data source.
Also every \textbf{OutgoingEvent} instance, i.e., every
diagnostic event, \textbf{relatesTo} some instance of
\textbf{DataSource}, because the end user must be able to
reconstruct which data source(s) provided information that caused
diagnostic rules to fire a given diagnostic event.
\textbf{OutgoingEvent} specializes to \textbf{AlarmEvent}, \textbf{FaultEvent}
and \textbf{DescriptorEvent}. Every \textbf{OutgoingEvent} instance is
connected to one of \textbf{DiagnosticIndicator} instances, 
i.e. \textbf{Alarm}, \textbf{Fault} and \textbf{Descriptor} sub-concepts,
 by \textbf{reports} relation, in order to have a reference message
 about the diagnostic rules.

\section{Computer-automated design of elevators}
\label{sec:elevator}
\begin{figure}[t!]
\begin{center}
\scalebox{.22}{\includegraphics{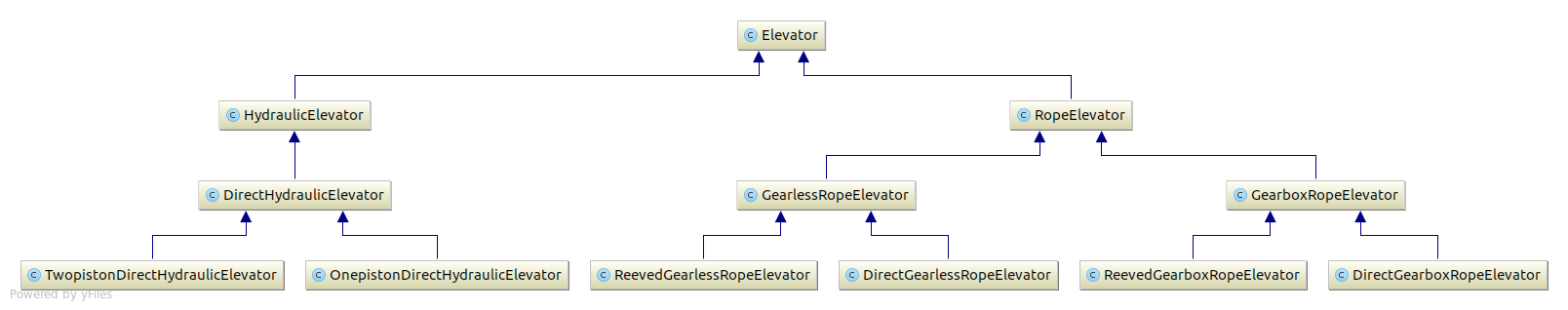}}\\
\scalebox{.26}{\includegraphics{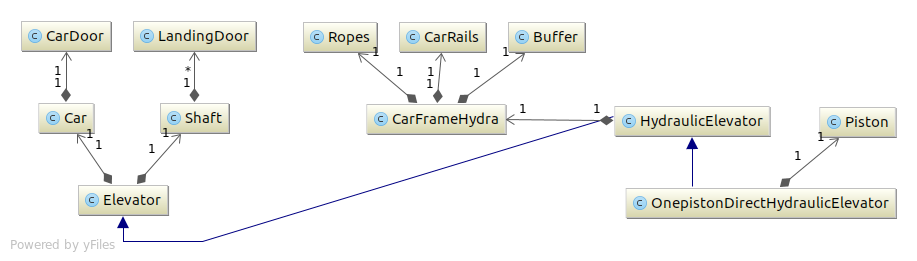}}
\end{center}
\caption{\label{fig:elev_onto} Ontologies describing the
  implements handled by \liftcreate{} (top) and 
  the components of \textsf{OnePistonHydraulicElevator}
  (bottom). Concepts are rectangles, concept inclusion is denoted by
  solid arrows, and HAS-A object properties are denoted by diamond-based arrows.}  
\end{figure}

Our latest ontology-based application is in the field of
computer-automated design (CautoD) which differs
from ``classical'' computer-aided design (CAD) in that it
is oriented to replace some of the designer's capabilities and not just
to support a traditional work-flow with computer graphics
and storage capabilities. Nevertheless, CautoD programs most
often include CAD facilities to visualize technical drawings related
to the implements of interest. 
In particular, our \liftcreate{} program is oriented to automating
design of elevators, taking the designer from the very first
measurements to a complete project which guarantees feasibility within
a specific normative framework. \liftcreate{} works in three steps. In
the first step,  the user is asked to enter relevant parameters
characterizing the project, and an overall ``design philosophy'' to be
implemented. For instance, if the size of the elevator's shaft is
known and fixed in advance, \liftcreate{} can generate solutions which
maximize payload,  door size, or car size. A design philosophy is just
a set of heuristics which, e.g., prioritize door size over other
elements, still keeping into account hard constraints, e.g., payload
and car size should not fall below some threshold. In the second
phase, \liftcreate{} retrieves components from a database of parts and
explores the (combinatorial) space of potential solutions, either
using heuristic search techniques, or resorting to optimizations
techniques --- like those suggested, e.g., in~\cite{ByeOPHS16}. In the
third phase, a set of feasible designs is proposed to the user, sorted
according to decreasing relevance considering the initial design
philosophy. For instance, if door size is to be maximized, the first
alternatives shown to the user are those with the widest doors,
perhaps at the expense of payload or car size.

The main issue with \liftcreate{} work-flow is that even simple
versions of elevators consists of a large number of components,
including car frame, car, doors (car and landing doors), emergency
brakes, pistons or cables, motors and control logic. In order to
explore the space of potential designs, components cannot be
solely available as drawing elements, like in classical CAD solutions,
but they must be handled as first class data inside \liftcreate{}
logic. This aspect required us to organize the taxonomy related to
different kinds of elevators and, for each elevator kind, to structure
the components in a part-whole hierarchy. In
Figure~\ref{fig:elev_onto} we show a fragment of the taxonomy for
elevators and an example of part-whole structure for a specific
elevator kind. In particular, in
Figure~\ref{fig:elev_onto} (top), we see that \liftcreate{} classifies
\textbf{Elevator} individuals in two main subclasses corresponding to
hydraulic-based (\textbf{HydraulicElevator}) and rope-based
(\textbf{RopeElevator}) designs. Both subclasses feature additional
partitions to handle specific design requirements, e.g., rope
elevators can be provided with a reduction gearbox or not, and the
drive can be direct of reeved. For one leaf class of the taxonomy,
namely \textbf{OnePistonDirectHydraulicElevator}, in
Figure~\ref{fig:elev_onto} (bottom) we show the detailed part-whole
diagram, from which we learn that, e.g., the only peculiar aspects of
such subclass is to have only one \textbf{Piston}, whereas the
remaining components are common to \textbf{HydraulicElevator} or
\textbf{Elevator}. Also we can see that the car frame is specific of hydraulic elevators (\textbf{CarFrameHydra}) and it is comprised of several parts, including
\textbf{CarRails}, \textbf{Buffer} and \textbf{Ropes}. The
relationships encoded in such part-whole hierarchy are
instrumental to \liftcreate{} when it comes to handle drawing, storage
and retrieval of designs, but also to reason about the various
trade-offs of a design when searching in the space of potential solutions.

\section{Conclusions}
\label{sec:concl}
Considering the experiences herein outlined, we summarize some
lessons learned in applying ontologies for systems engineering. First
and foremost, while ontologies provide an effective tool for
conceptualizing scenarios as diverse as those considered, 
some ontology-based tools, e.g., DL reasoners, are untenable
unless small-to-medium scale systems are considered. In the case of
E414 ontology reasoning with an expressive ontology required us to
implement strategies to ``forget'' data to avoid cluttering
the reasoner. In the ILS ontology, where SPARQL queries for KPIs are
the only reasoning requested and the usage of OWL 2 QL profile banned
expressive but hard-to-compute constructs, scaling 
still requires discarding data using a recency approach. On the other hand, 
in DiSeGnO and \liftcreate{}, ontologies merely provide means for
conceptualizing data and, as such, flexibility is gained without
sacrificing performances. The second take-home message is that
sublanguages of OWL 2 are adequate for most modeling purposes. 
With the only exception of E414 ontology, the ones herein
considered fit OWL 2 QL constraints which allowed us
to combine in a  natural way subclassing (``IS-A'' relationships) with 
other kind of object properties (including ``HAS-A''). However, the fact that OWL 2 QL  
ontologies can be compiled to relational databases --- as in the case
of DiSeGnO --- or handled trough an object-persistency module --- as
in the case of \liftcreate{} --- makes their use transparent to other
system components. Third, and final point, with the exception of ILS
monitoring, none of our applications required the integration of
different data sources which is indeed one of the main tasks which
ontologies are advocated for. Nevertheless, our experience witnesses
that even in single-source data modeling, ontologies provide an
excellent mean to bridge the gap between domain experts and
computer software designers.




\bibliographystyle{unsrt}

%









\end{document}